\documentclass[10pt,twocolumn,letterpaper]{article}

\usepackage[pagenumbers]{cvpr} 

\usepackage{graphicx}
\usepackage{amsmath}
\usepackage{amssymb}
\usepackage{booktabs}
\usepackage{adjustbox}
\usepackage{multirow}
\usepackage[accsupp]{axessibility}

\usepackage{hyperref}
\hypersetup{breaklinks,colorlinks}

\usepackage[capitalize]{cleveref}
\crefname{section}{Sec.}{Secs.}
\Crefname{section}{Section}{Sections}
\Crefname{table}{Table}{Tables}
\crefname{table}{Tab.}{Tabs.}

\def\cvprPaperID{9306} 
\def\confName{CVPR}
\def\confYear{2023}

\begin{document}

\title{MoStGAN-V: Video Generation with Temporal Motion Styles}

\author{Xiaoqian Shen \qquad Xiang Li \qquad Mohamed Elhoseiny \\
King Abdullah University of Science and Technology (KAUST) \\
{\tt\small \{xiaoqian.shen, xiang.li.1, mohamed.elhoseiny\}@kaust.edu.sa}
}

\maketitle

\begin{abstract}

Video generation remains a challenging task due to spatiotemporal complexity and the requirement of synthesizing diverse motions with temporal consistency. 
Previous works attempt to generate videos in arbitrary lengths either in an autoregressive manner or regarding time as a continuous signal. However,  they struggle to synthesize detailed and diverse motions with temporal coherence and tend to generate repetitive scenes after a few time steps.
In this work, we argue that a single time-agnostic latent vector of style-based generator is insufficient to model various and temporally-consistent motions. Hence, we introduce additional time-dependent motion styles to model diverse motion patterns.
In addition, a \textbf{Mo}tion \textbf{St}yle \textbf{Att}ention modulation mechanism, dubbed as MoStAtt, is proposed to augment frames with vivid dynamics for each specific scale (i.e., layer), which assigns attention score for each motion style w.r.t deconvolution filter weights in the target synthesis layer and softly attends different motion styles for weight modulation. Experimental results show our model achieves state-of-the-art performance on four unconditional $256^2$ video synthesis benchmarks trained with only 3 frames per clip and produces better qualitative results with respect to dynamic motions. { Code and videos have been made available at \url{https://github.com/xiaoqian-shen/MoStGAN-V}}.
\end{abstract}


\section{Introduction}
\label{sec:intro}
Learning algorithms for image generation are rapidly reaching maturity and are expected to soon approach human levels of performance. However, the video generation task does not share similar success and remains a research opportunity to be  further explored.
Since videos are computationally intensive to model, a key question is how to generate high-quality videos with limited computation resources. 
Another challenge is that due to the very nature of video data, video synthesis is not simply tantamount to generating high-quality images that change over time but also requires generating these frames with temporal consistency, i.e., natural transitions between frames as well as realistic motions. This problem gets even more severe when the number of elements (e.g., objects and background) that move spatiotemporally grows.

\begin{figure}[t!]
\begin{adjustbox}{width=\linewidth,center}
    \centering
    \includegraphics{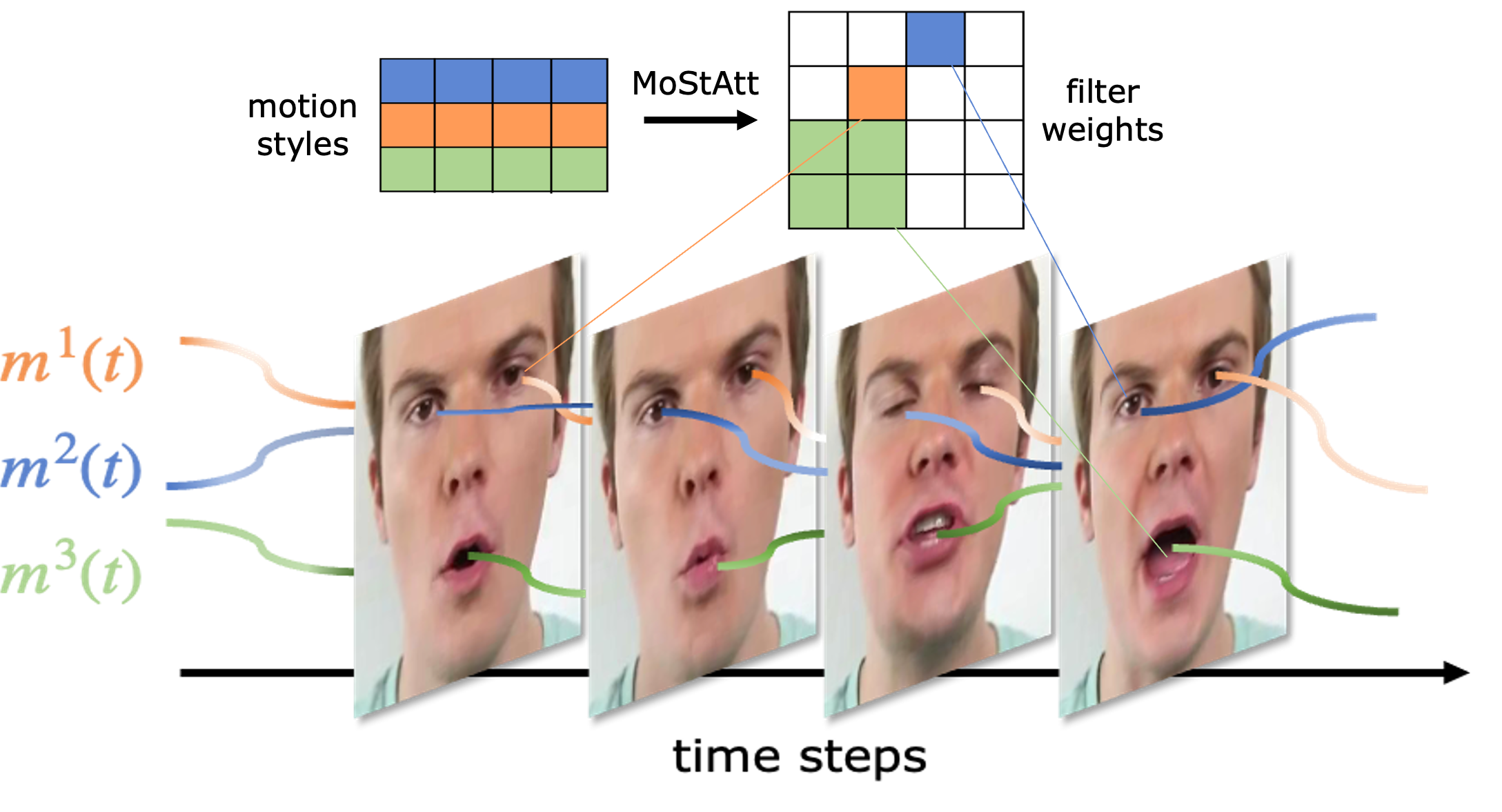}
\end{adjustbox}
\caption{Intuitively show how motion styles model different motion patterns along time steps. Each curve corresponds to one motion style as a function of time step $t$. Each motion style will adaptively modulate filter weights of the generator with our proposed MoStAtt mechanism and further stylize input features with dynamic motions.}
\end{figure}

Towards low computation consumption, several RNN or LSTM-based autoregressive approaches~\cite{tulyakov2018mocogan,clark2019adversarial,saito2020train,tian2021mocogan} reduce computation complexity and make long video generation possible but the resulting videos tend to accumulate errors over time and contain inconsistent frames.
Recent works use implicit neural representations by regarding time as continuous signals and mapping time to either spatial coordinates~\cite{yu2022digan} or StyleGAN~\cite{karras2020training} feature inputs~\cite{skorokhodov2022stylegan}. Although these methods leverage sparse training to enhance training efficiency and produce arbitrarily long videos, they fall short in capturing diverse patterns of changing motions. For instance, a generated video of a person talking may contain only low-frequency global motions like moving head to different angles but lack the ability to synthesize high-frequency mouth and eye actions, e.g., open/close mouth and blink eyes. 

The limitation of previous works leads us to rethink how to explicitly model the diversity of motions and leverage them for motion synthesis in the video generation process.
Motion segmentation methods~\cite{elqursh2013online,keuper2018motion,bideau2018moa} assume similar trajectories representing for similar motions and segment videos into static background and diverse moving objects, which
explicitly model them on perception level. Drawing inspiration from this perceptual grouping concept commonly employed in motion segmentation literature, we formulate the notion of diverse motion modeling for video generation on top the style-based framework ~\cite{karras2020training,karras2020analyzing,skorokhodov2022stylegan} which leverage style parameters to control different levels of synthesis by modulating the weights of scale-specific layers in the generator. This means we aim to design group of motion styles, each of which corresponds to one motion pattern and is able to stylize input features with dynamic motions by adaptively modulating the weights of deconvolutional filters.

In this work, we aim at  generating videos with vivid dynamics and rich motions over time while reserving temporal consistency. We argue that single time-agnostic latent vector of the StyleGAN-based model is insufficient to model different motion patterns in complicated motion-variant video datasets and consider auxiliary style control to increase motion awareness in the generation process. To this end,  we introduce the concept of motion styles 
for video generation and develop a motion network to generate time-dependent motion styles to model diverse motion patterns along continuous time. In addition, we consider how to make multiple motion styles dynamically contribute to motion synthesis since motions might temporally exist in several frames while disappear in other time steps.
Therefore, a motion-style attention modulation mechanism, dubbed as MoStAtt, is designed to 
perform cross-attention between deconvolutional filters and motions styles and linearly combine filter-specific and time-dependent motion styles for dynamic filter weight modulation. Variant motion styles that correspond to different motion patterns will adaptively attend for frames stylization and thus the modulated weights can better represent time-dependent and diverse motions. Note that our approach differs from previous works, which either predict latent motion trajectories for an image generator~\cite{tian2021mocogan} or generate continuous motion codes and concatenate them with constant vectors to serve as initial feature inputs for a generator network~\cite{skorokhodov2022stylegan}. In contrast, our proposed motion styles and MoStAtt modules are designed to modulate kernel weights, rather than functioning as feature inputs.



Our contributions are highlighted below:

\begin{itemize}
\item Introducing time-dependent motion styles for video generation task in addition to the original time-agnostic content style of style-based models to facilitate weight modulation and thus raise temporal awareness and enhance motion synthesis. 

\item Motion style attention modulation mechanism, dubbed as MoStAtt, which softly attends different motion styles for weight modulation in each synthesis layer and facilitates the modulated weights to augment individual static frames with dynamic motions across holistic time continuous videos.

\item Our simple yet effective approach achieves state-of-the-art performance in unconditional video generation and enhances qualitative results of motion synthesis.
\end{itemize}

\section{Related Work}
\label{sec:related_work}

\subsection{Video Synthesis}
Different from image generation, which only models pixels at the spatial level, generating video from scratch is a more challenging task since it takes an additional temporal dimension into consideration. VGAN~\cite{vondrick2016generating} adapts GAN to generate foreground scenes using 3D deconvolution and combines 2D background to create videos with a mask. MoCoGAN~\cite{tulyakov2018mocogan} and TGAN~\cite{saito2017temporal} model spatial and temporal dimensions with an image generator and a RNN model separately. MoCoGAN-HD~\cite{tian2021mocogan} proposes to predict a sequence of latent motion trajectory by training a motion generator and fed it into pre-trained StyleGAN to synthesize a sequence of images. StyleGAN-V~\cite{skorokhodov2022stylegan} regards time as continuous signals while DIGAN~\cite{yu2022digan} builds on top of INR-GAN~\cite{skorokhodov2021adversarial} and treats videos as continuous signals at both spatial and temporal dimensions. LongVideoGAN~\cite{brooks2022generating} leverages a hierarchical generator in a multi-scale training strategy to synthesis first-person viewpoint videos. Our method builds on top of style-based model and considers time as a continuous signal similar to~\cite{skorokhodov2022stylegan,yu2022digan} and extends style-based models with additional temporal dependent motion styles and prioritizes motion awareness with newly proposed style-based attention weight modulation technique.

In addition, several works compress high dimensional video data into a discretized latent space. For instance, VideoGPT~\cite{yan2021videogpt} utilizes VQ-VAE~\cite{van2017neural} to encode video data to latent sequences for the prior model to predict target sequences. TATS~\cite{ge2022long} extends VQ-GAN~\cite{esser2021taming} to 3D and conditions one interpolation transformer on a sparse autoregressive transformer to predict tokens. Recently, diffusion models have also been applied in the video generation task~\cite{ho2022video,yang2022diffusion,hong2022cogvideo}. But they require huge computation resources, with millions of time steps to achieve high quality results and also suffer from slow inference speed. 
For example, a single frame of video generated with Disco Diffusion~\cite{katherine2021disco} takes on the order of 5 minutes and 17 seconds for animation adaptations of Stable Diffusion~\cite{rombach2022high}, and 1 minute for CogVideo~\cite{hong2022cogvideo}. However, our method upholds a computation effective manner that trains model with only 3 frames for each video clip and is able to generate videos at 3.12 millisecond per frame during inference time.

\subsection{HyperNetworks}

HyperNetworks refer to the models that use an auxiliary light network to generate parameters for the main networks~\cite{ha2016hypernetworks,littwin2020optimization}. Hypernetworks have been proved useful in many fields like few-shot learning~\cite{bertinetto2016learning}, continual learning~\cite{von2019continual}, and language modeling~\cite{suarez2017language}. In the case of generative models, \cite{anonymous2023adversarial} utilizes HyperNetworks to transform text-conditional signal to modulate weights of the generator for text-controllable image generation. In this work, we leverage a HyperNet-augmented modulation approach to facilitate motion synthesis in video generation task, which means weight modulation matrix of each deconvolutional layer are produced with a HyperNetwork. HyperNetworks suffer from extreme memory consumption due to additional parametrization. To alleviate this issue, we adopt a low-rank modulation technique to obtain the modulation weight matrix as the product of factorized modulating tensors produced by a HyperNetwork to improve training efficiency.

\subsection{Latent Space Decomposition}
 Style-based models~\cite{karras2020training,karras2020analyzing,skorokhodov2022stylegan} transform latent vectors to different modulation based convolutional layers for fine-grained synthesis control. Alterations of the latent code correspond to particular manipulations in generated images. Recent works~\cite{wu2021stylespace,shen2021closed} attempt to figure out the valid directions in the high-dimensional latent space for interpretable style control. \cite{wang2022latent} decompose motion codes into orthogonal basis deviating from source images for image animation. \cite{qiu2022stylefacev} decomposes images into appearance and pose representations and re-compose both representations to obtain latent code $w$ for facial video generation. Our method separates a new branch to generate time-dependent motion styles in parallel with original content vector of style-based models and separately modulate the weights of deconvolutional layers with both time-agnostic content style as global context and time-sensitive motion styles to augment static frames with dynamic motions.
 

\section{Method}

\begin{figure*}[!htbp]
\begin{adjustbox}{width=\linewidth,center}
    \centering
    \includegraphics{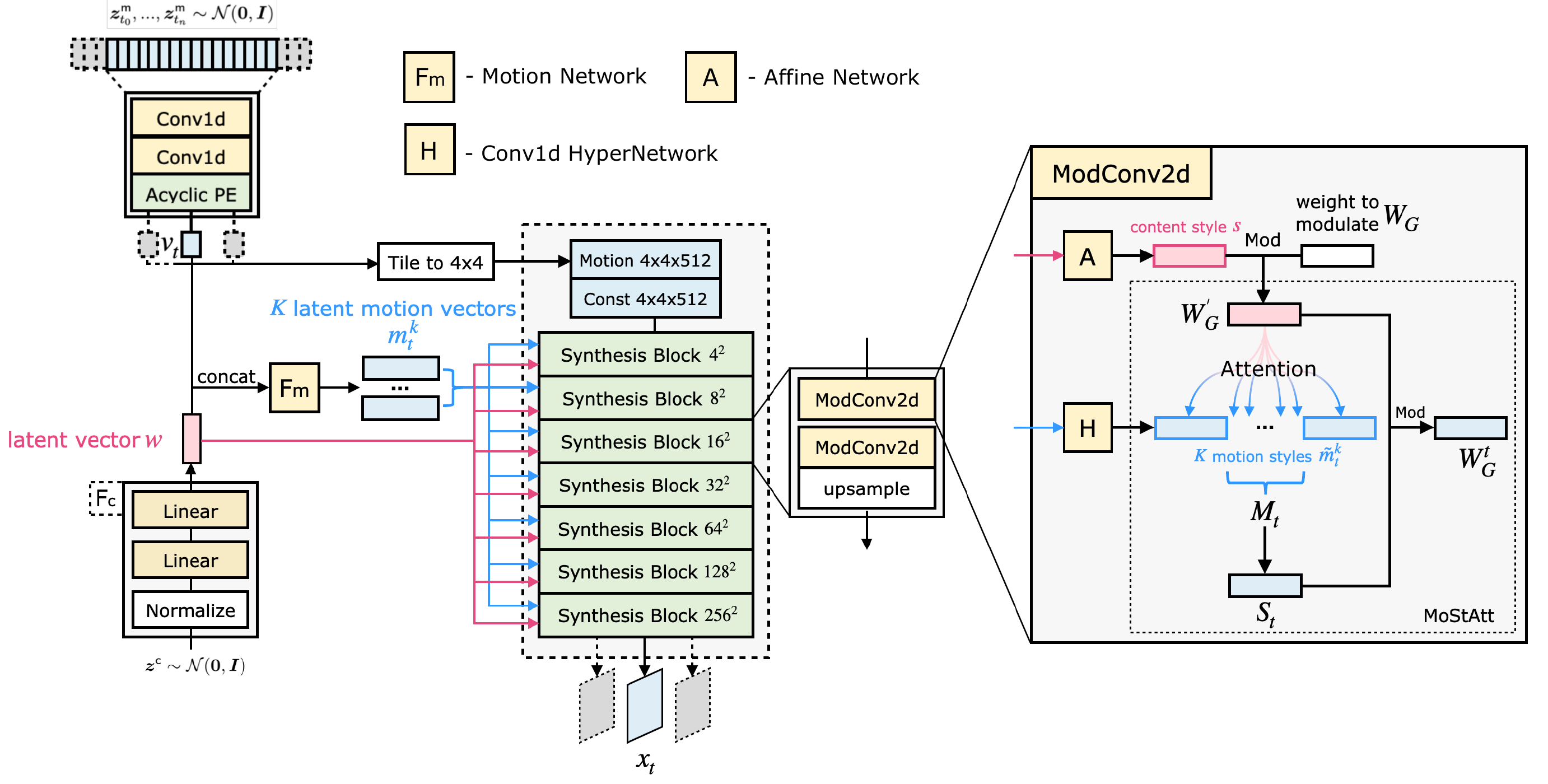}
\end{adjustbox}
\caption{Overview of MoStGAN-V generator.  The Motion Network $\sf{F_m}$ takes the concatenation of motion code $v_t$ and latent vector $w$ as input to generate time-dependent $K$ latent motion vectors $\{m_t^k\}_{k=1,2,...,K}$ for each time step $t$. The latent vector $w$ and $K$ latent motion vectors will be transformed by Affine Network $\sf{A}$ and HyperNet $\sf{H}$ into content style $s$ and motion styles $\{\tilde{m}_t^k\}_{k=1,2,...,K}$ respectively to control the weights of each Synthesis Block. Within \texttt{ModConv2d}, the content style first modulate the weight $\mathbf{W}_G$ of deconvolutional filter same as vanilla style-based model~\cite{karras2020training}. Then MoStAtt mechanism first performs cross attention between weight matrix $\mathbf{W}_G^{'}$ and modulation matrix $\mathbf{M}_t$ (consists of $K$ motion styles) and softly attends different motion styles as motion modulation matrix $\mathbf{S}_t$  And then the weight matrix $\mathbf{W}_G^{'}$ will be modulated by $\mathbf{S}_t$, resulting in the final modulated weights $\mathbf{W}_G^t$ of deconvolution layer for time step t. This mechanism will be elaborated in Section~\ref{sec:attn}.}
\label{fig:main}
\end{figure*}

\subsection{Method Overview}
\textbf{Generator.}
Our generator derives from StyleGAN-V~\cite{skorokhodov2022stylegan}, which treats time as a continuous signal and can generate arbitrary length of frames. It models motion noise $z_{t_0}^m,...,z_{t_n}^m \in \mathcal N(0, I)$ with \texttt{Conv1d} layers, where $m$ represents for motion noise and $t_0,...,t_n$ denote discrete time steps in chronological order. Then it uses acyclic positional encoding to predicts amplitudes, periods, and phases of corresponding waves for motions of different frequencies to obtain motion code $v_t$ for current time step $t$ (see left-top part of Figure~\ref{fig:main}). Then $v_t$ is concatenated  with trainable constant vector as generator inputs.

Same as style-based methods~\cite{karras2020training,karras2020analyzing,skorokhodov2022stylegan}, our generator samples a noise vector $z^c \in \mathcal N(0, I)$ from a Gaussian distribution and passes it through a Mapping Network $\sf{F_c}$ to get latent vector $w \in \mathbb{R}^{d_c}$ in intermediate latent space $\mathcal{W}$~\cite{abdal2019image2stylegan}. For each synthesis layer, $w$ will be transformed by an Affine Network $\sf{A}$ to generate content style $s \in \mathbb{R}^{c_{in}}$ in $\mathcal{S}$ space~\cite{wu2021stylespace}. The content style then modulates the filter weight $\mathbf{W}_G \in \mathbb{R}^{c_{out} \times c_{in}\times k_h \times k_w}$ of corresponding convolutional layer in the generator by element-wise product, i.e., $\mathbf{W}_G^{'} = \mathbf{W}_G \odot \mathbf{W}_c$, where $\mathbf{W}_c$ is obtained by broadcasting $s$ to $\mathbb{R}^{c_{in} \times k_h \times k_w}$, and $c_{out}, c_{in}, k_{h}, k_{w}$ represent for output channel, input channel, and kernel size of current filter respectively.

In this study, instead of using motion codes only as generator inputs, we argue that time-dependent motion styles can be generated from motion codes and further used to modulate the weights of synthesis layers for diverse motion synthesis. Figure \ref{fig:main} gives an overview of the generator network. We elaborate on the generation process of time-sensitive motion styles in Section~\ref{sec:motion_style} and propose a new attention-based modulation mechanism MoStAtt in Section~\ref{sec:attn}.

\textbf{Discriminator.} For computation efficiency, we follow previous work~\cite{skorokhodov2022stylegan} that uses a 2D discriminator network instead of 3D convolutional networks to independently extract features for each individual frame. After reaching a lower resolution block of the discriminator, it concatenates all the frame features within a video by time dimension as the global representation for the whole video. At the last block, the discriminator computes time distance information encoded with positional encodings and outputs real/fake logits as a dot product between time difference embeddings and the global feature vector. 

However, we think such an operation still could not enable the discriminator to capture the motion differences in feature level, thus, we also feed frame differences as auxiliary discriminator inputs to facilitate motion artifact capture, which will be elaborated in Section.~\ref{sec:motion_diff}.

\subsection{Time-sensitive Motion Styles}
\label{sec:motion_style}

Although StyleGAN-V~\cite{skorokhodov2022stylegan} represents time as a continuous signal and concatenates the time-dependent motion code $v_t$ on top of the constant vector as generator inputs; however, it struggles to model motions of different patterns, which means the dynamic motion and the relatively static background will have a similar frequency to move, (i.e., the head might resemble a similar moving trajectory as the mouth in a talking head generation case). The main reason is that the content style mentioned above is invariant to time and lacks temporal information, thus, it is insufficient to model the whole video, which contains complex motions of variant motion patterns. To address this issue, \emph{we augment the generation process with additional time-dependent motion styles apart from the original content style by leveraging two types of styles to separately modulate weights of each synthesis layer of the generator}.

In original StyleGAN-V~\cite{skorokhodov2022stylegan} method, each motion code $v_t$ is generated from a sequence of trajectory noise and processed by several $\texttt{conv1d}$ layers with a large filter (e.g., kernel size is 11), it, therefore, has already received long-term temporal information across different time steps. In this work, we want to utilize such knowledge to generate time-dependent motion styles. More specifically, the latent vector $w$ will be concatenated with motion code $v_t \in \mathbb{R}^{d_v}$ and then pass through a Motion Network $\sf{F_m}$ to generate $K$ time-dependent motion vectors $\{m_t^k\}_{k=1,2,...,K} \subset \mathbb{R}^{d_m}$.
A HyperNetwork is then designed to transform these motion vectors into a weight modulation matrix to modulate the weights in each synthesis layer. However, a HyperNetwork that directly produces a full rank matrix of dimensionality $c_{in}\times c_{out} \times k_h \times k_w$ for each deconvolutional filter will consume extremely large memory. To address memory-intensive hyperscaling, we adopt a low-rank tensor decomposition technique to improve training efficiency~\cite{suarez2017language}.
More specifically, our HyperNetwork $\sf{H}$ transforms motion vectors $\{m_t^k\}_{k=1,2,...,K}$ into motion styles $\{\tilde{m}_t^k\}_{k=1,2,...,K} \subset \mathbb{R}^{R\times (c_{in} + k_h + k_w)}$ for each synthesis layer to modulate the deconvolution weights, where $R$ denotes the rank number in tensor decomposition process. The full rank motion modulation matrix for each motion style ${\tilde{m}_t^k}$ can be calculated as:
\begin{equation}
    \mathbf{M}_t^k = \sum_{r=1}^R v_1^r \otimes v_2^r \otimes v_3^r
\label{eq:rank}
\end{equation}
where $\mathbf{M}_t^k \in \mathbb{R}^{c_{in} \times k_h \times k_w}$ and $\otimes$ denotes outer product. $v_1^r$, $v_2^r$, and $v_3^r$ denote modulating vectors slicing from ${\tilde{m}_t^k}$, with sizes of $ c_{in}, k_h, k_w$ respectively. Eventually, the whole modulation matrix for $K$ motion styles is $\mathbf{M}_t \in \mathbb{R}^{K\times (c_{in} \times k_h \times k_w)}$.

\subsection{MoStAtt: Motion Style Attention Modulation}
\label{sec:attn}

Inspired by ~\cite{anonymous2023adversarial} which utilizes text-conditional signal for weight modulation and leverage attention mechanism for text-controllable image generation, we explore attention mechanisms for motion style weight modulation in the case of unconditional video generation and aim to make motion styles softly attend modulation operation for diverse motion pattern modeling.
Representing videos as a combination of individual frames, we integrate $t$ into batch dimension and perform attention between the modulation matrix $\mathbf{M}_t$ and deconvolutional filter weights in each synthesis layer formulated as below:
\begin{equation}
\mathbf{S_t}=\texttt{Softmax}(\frac {\mathbf{W}_G^{'}(\mathbf{M}_t)^T} {\sqrt{c_{in}\times {k_h}\times {k_w}}})\mathbf{M}_t 
\label{eq:attn1}
\end{equation}
where $\mathbf{S}_t \in \mathbb{R}^{c_{out} \times (c_{in} \times k_h \times k_w)}$ denotes the final motion modulation matrix. Intuitively, this operation determines which motions should attend in current frame, given the whole context. Therefore, the motion-attended weights will decorate the generated frames with diverse motions and progressively interact with contextualized information layer by layer via MoStAtt mechanism. 
The modulated weights for current frame will be finalized as $\mathbf{W}_G^t = {\mathbf{W}_G^{'}} \odot \mathbf{S}_t $, thus the modulated weights of each $\texttt{ModConv2d}$ layer will be aware of temporal information from $\mathbf{S}_t$. Please check the right part of Figure \ref{fig:main} for illustration.

\subsection{Motion Diversity}
\label{sec:motion_div}

To encourage motion styles to be disentangled from each other for modeling various motions patterns, we further leverage orthogonality constraints introduced in ~\cite{salzmann2010factorized} to prevent motion modulation weights from overlapping. More specifically, we obtain the attention matrix $\mathbf{A_t}=\frac {\mathbf{W}_G^{'}(\mathbf{M}_t)^T} {\sqrt{c_{in}\times {k_h}\times {k_w}}}$ (from Eq.~\ref{eq:attn1}) before softmax layer and calculate a regularization loss for each \texttt{ModConv2d} as follows:
\begin{equation}
    L_{div} =\frac{1}{T} \sum_{t=1}^T ||\mathbf{A_t}^T \mathbf{A_t}||_{F}
\end{equation}
where $||\cdot||_{F}$ is the Frobenius norm and we average over all \texttt{ModConv2d} layers to form a motion regularization loss.

\begin{table*}[!htbp]
    \centering
\begin{adjustbox}{width=\linewidth,center}
\begin{tabular}{cccccccccc}
\toprule  \multirow{2}{*}{ \textbf{Method} } & \multicolumn{2}{c}{ \textbf{FaceForensics $\mathbf{256^2}$} } & \multicolumn{2}{c}{ \textbf{SkyTimelapse $\mathbf{256^2}$} } & \multicolumn{2}{c}{ \textbf{RainbowJelly $\mathbf{256^2}$} } & \multicolumn{2}{c}{ \textbf{CelebV-HQ $\mathbf{256^2}$} }
 \cr & $\rm{FVD_{16}}$ & $\rm{FVD_{128}}$ & $\rm{FVD_{16}}$ & $\rm{FVD_{128}}$ & $\rm{FVD_{16}}$ & $\rm{FVD_{128}}$ & $\rm{FVD_{16}}$ & $\rm{FVD_{128}}$\\ 
 \midrule
 MoCoGAN-HD~\cite{tian2021mocogan} &111.8 &653.0 &164.1&878.1&579.1&628.2&212.4&753.1 \\
VideoGPT~\cite{yan2021videogpt} &185.9 & N/A &222.7&N/A&136.0&N/A&177.8&N/A \\
DIGAN~\cite{yu2022digan} &62.5 &1824.7 &83.1&196.7&436.6&369.0&72.9&163.2 \\
StyleGAN-V~\cite{skorokhodov2022stylegan} &47.4&89.3&79.5&197.0&195.4&262.5&68.0&158.6 \\
MoStGAN-V (ours) & \textbf{39.7} & \textbf{72.6} & \textbf{65.3} & \textbf{162.4} & \textbf{70.1} & \textbf{74.3} & \textbf{56.1} & \textbf{132.1} \\

\bottomrule
\end{tabular}
\end{adjustbox}
\caption{Comparison of quantitative performance among unconditional video generation models. For each method, we report the result with the best $\rm{FVD_{16}}$ performance.}
\label{tab:main}
\end{table*}

\subsection{Motion Consistency}
\label{sec:motion_diff}
Another issue that hinders the performance of existing video generation methods is that they fail to preserve long-term consistency, e.g., motions will become receptive after several frames. To alleviate periodic artifacts, we strengthen the discriminator with frame discrepancy to capture motion changes among different frames. More specifically, we compute differences between every frame pair in a video and concatenate them with either generated or real videos as the discriminator's input. Therefore, the input of the discriminator will be: 
\begin{equation}
(x_{1},..., x_{N}, \delta x_1, ...,\delta x_{N-1}), \mathop{\delta x_{i}}\limits_{1\le i \le N-1}:=|x_{i+1}-x_{i}|
\end{equation}
where $(x_{1},..., x_{N)}$ represent for $N$ generated frames for a video clip. Intuitively, since consecutive frames are correlated due to their time continuity, this operation encourages the discriminator to focus on motion artifacts among frames. 

Note that StyleGAN-V~\cite{skorokhodov2022stylegan} also attempted in a similar way that it computes
differences between activations of next/previous frames in a video and concatenates this difference to the original activation maps
, but it did not work because the discriminator is far more powerful and outpaces the generator much. In contrast, we observed that it works for our approach since our layer-wise MoStAtt augmented generator can produce realistic enough motions and fight against the discriminator, making it more capable of capturing motion artifacts.

\section{Experiments}


\subsection{Experimental Setups}

\noindent \textbf{Datasets.} Our MoStGAN-V model is evaluated on 4 unconditional benchmarks: FaceForensics $256^2$~\cite{rossler2018faceforensics}, SkyTimelapse $256^2$~\cite{xiong2018learning}, RainbowJelly $256^2$ \cite{skorokhodov2022stylegan} and CelebV-HQ~$256^2$\cite{zhu2022celebv}.
We provide the details of the dataset in supplementary.

\noindent \textbf{Evaluation metrics.} Following previous works, we use Frechet Video Distance (FVD)~\cite{unterthiner2018towards} to evaluate the video generation performance. However, FVD is heavily dependent on data preprocessing procedures and sensitive to different sampling strategies (Appendix C in~\cite{skorokhodov2022stylegan}). Following~\cite{skorokhodov2022stylegan}, we prepare 2048 frames from real video and subsample the generated video into 16-frame and 128-frame segments, respectively, denoted as $\rm{FVD_{16}}$ and $\rm{FVD_{128}}$, to evaluate the generation quality. For all the experiments we select the results with the best $\rm{FVD_{16}}$ performance. In addition, we also conduct a human evaluation with respect to motion diversity and time consistency.

\noindent \textbf{Comparison Approaches.}

\textbf{- MoCoGAN-HD}~\cite{tian2021mocogan} decomposes content and motion synthesis and introduces a motion generator to discover the desired trajectory in the latent space of a pre-trained image generator.

\textbf{- VideoGPT}~\cite{yan2021videogpt}
first trains a VQ-VAE~\cite{van2017neural} model for discrete latent representations of videos and then a GPT-like architecture to autoregressively model the discrete latents. 

\textbf{- DIGAN}~\cite{yu2022digan} utilizes an INR-based video generator to improve the motion dynamics by manipulating the space and time coordinates separately.

\textbf{- StyleGAN-V}~\cite{skorokhodov2022stylegan} regards time as continuous signals and concatenates the continuous motion codes on top of a constant vector in StyleGAN2~\cite{karras2020training} architecture.

\noindent \textbf{Implementation Details.}
All models are trained on 32GB NVIDIA V100 GPUs. Our model uses one node of 4 GPUs and takes less than 2 days to converge to the lowest $\rm{FVD_{16}}$. The proposed Motion Network $\sf{F_m}$ is implemented by several MLP layers which map the concatenation of motion code $v_t$ and latent vector $w$ into $K$ latent motion vectors $\{m_t^k\}_{k=1,2,...,K}$ with a dimension of 128, followed by a \texttt{conv1d}-based HyperNetwork with kernel size $1$ to transform them into motion styles $\{\tilde{m}_t^k\}_{k=1,2,...,K}$. We observe that increasing the rank in Eq. (\ref{eq:rank}) would not improve performance but increase parameters thus set rank $R$ to 1. By default, we set $K$ to 8 in our experiments. 

\begin{figure*}[!htbp]
\begin{adjustbox}{width=\linewidth,center}
    \centering
    \includegraphics{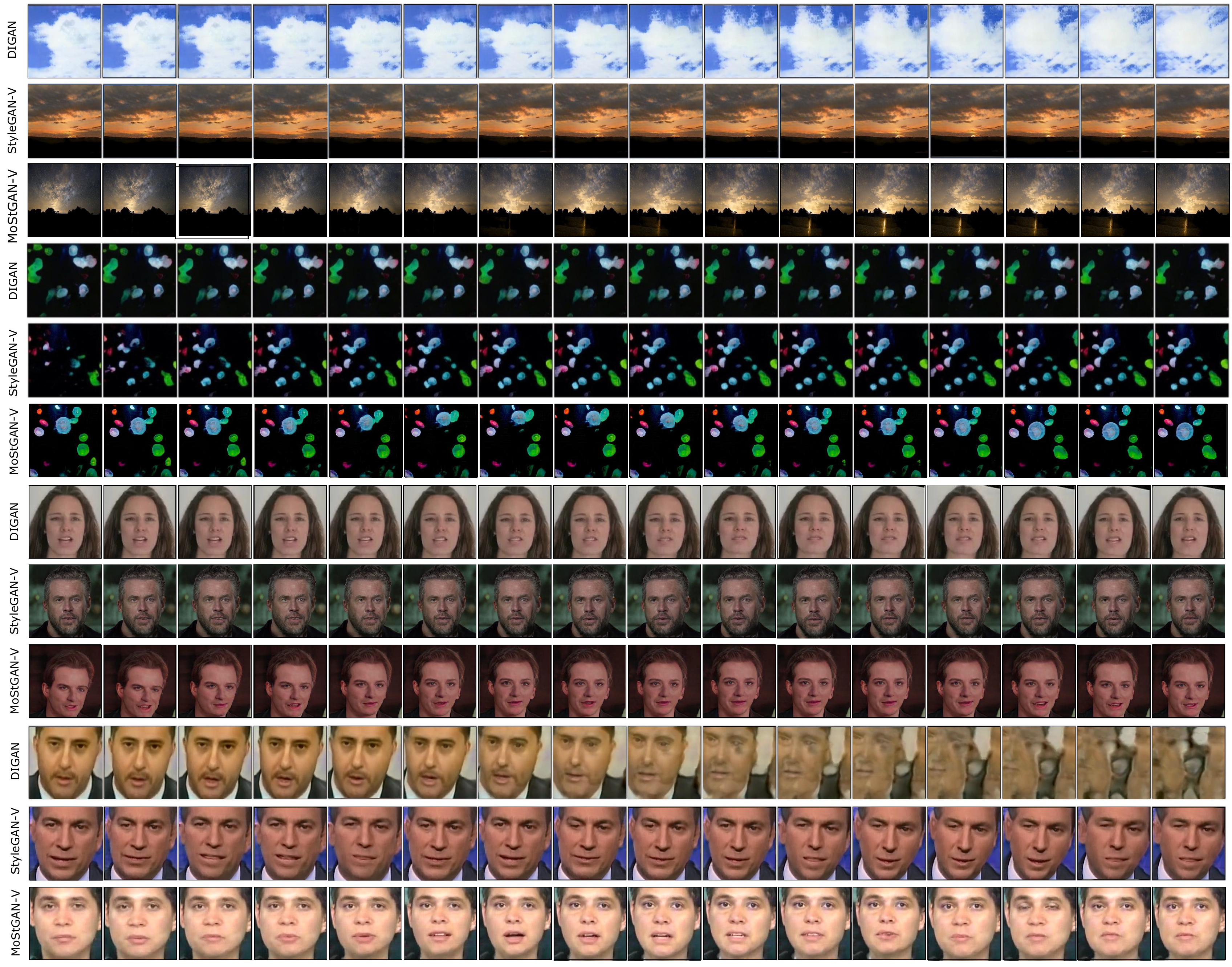}
\end{adjustbox}
\caption{Random samples from the comparison baselines and our model on SkyTimelapse $256^2$, RainbowJelly $256^2$, CeleV-HQ $256^2$ and FaceForensics $256^2$ respectively. Start from $t=0$ and report every 4-th frame from a 64-frame video clip. To better compare the qualitative results, please refer to synthesised videos in our project webpage.}
\label{fig:qual}
\end{figure*}

\subsection{Unconditional Video Generation}

\noindent \textbf{Quantitative Comparison.} As shown in Table~\ref{tab:main}, our MoStGAN-V achieves new state-of-the-art performance on all datasets. Generally, $\rm{FVD_{128}}$ score is higher than $\rm{FVD_{16}}$ since 128-frames-long videos require more coherent features to fit the real distribution. It is interesting to find that DIGAN~\cite{yu2022digan} reports a lower $\rm{FVD_{128}}$ score than $\rm{FVD_{16}}$ on the RainbowJelly dataset, while ours has a closer $\rm{FVD_{16}}$ and $\rm{FVD_{128}}$ score. One possible explanation is this dataset contains jellyfish moving back and forth, resulting in similar frames after several time steps. Note that the RainbowJelly dataset contains complex hierarchical motions, which makes this benchmark more challenging than others.
Our MoStGAN-V method successfully achieves the best performance on this motion-diverse dataset with the proposed MoStAtt mechanism.

\noindent \textbf{Qualitative Results.} Figure~\ref{fig:qual} shows qualitative results of unconditional video generation among five comparing methods. The major qualitative difference in results is that our model preponderates other approaches by generating realistic detailed motions assisted with motion style modulation. For instance, in CelebV-HQ $256^2$~\cite{zhu2022celebv} datasets, MoCoGAN-HD~\cite{tian2021mocogan} merely generates 
consecutive images without any consistent motions. VideoGPT~\cite{yan2021videogpt} and DIGAN~\cite{yu2022digan} pay attention to motion synthesis but fail to preserve quality in longer time steps. StyleGAN-V~\cite{skorokhodov2022stylegan} can only generate stable longer videos with moving faces but lacks of detailed motion synthesis. In contrast, our model succeeds in fine-grained and consistent facial expression and motion generation, (e.g. open/close for mouth and blinking eyes). Therefore, after introducing motion style and implementing MoStAtt mechanism on top of StyleGAN-V~\cite{skorokhodov2022stylegan}, people can actually talk!

\begin{figure}[!t]
\begin{adjustbox}{width=\linewidth,center}
    \centering
    \includegraphics{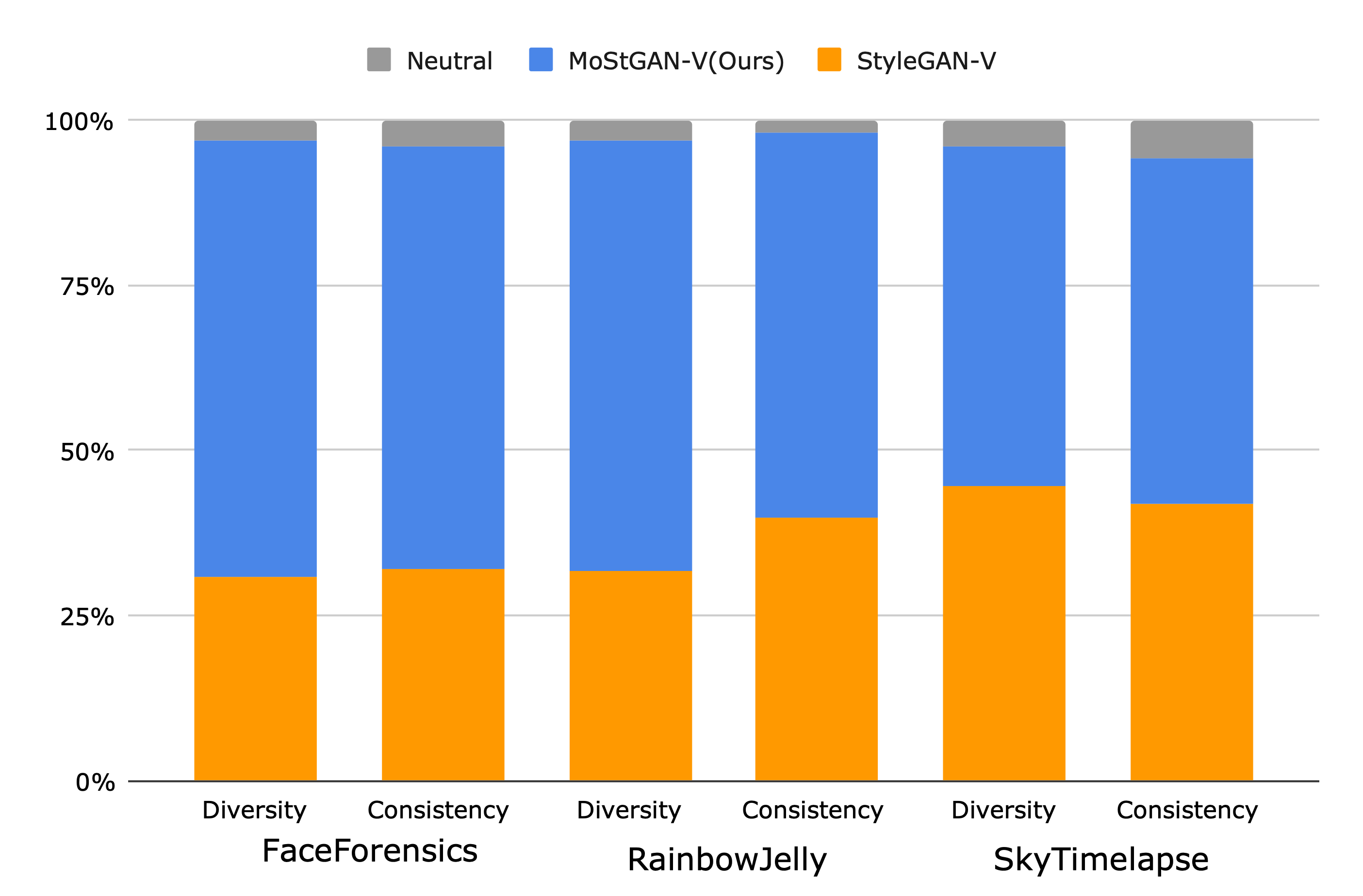}
\end{adjustbox}
\caption{Human evaluation results on FaceForensics $256^2$, RainbowJelly $256^2$, and SkyTimelapse $256^2$ datasets w.r.t motion diversity and temporal consistency.}
\label{fig:human}
\end{figure}

\noindent \textbf{Human Evaluation.}
In addition, we use Mechanical Turk to assess the quality of 100 generated videos per method for each dataset on both motion diversity and temporal consistency aspect. Given a pair of videos generated by StyleGAN-V~\cite{skorokhodov2022stylegan} and our MoStGAN-V models trained on the same dataset, people are asked to decide which video is better w.r.t motion diversity and time consistency separately. Each video is evaluated by 5 unique workers. We provide a neutral option if the volunteers find it's hard to decide which model is better. From Figure \ref{fig:human}, our MoStGAN-V model shows significantly better video generation quality on all evaluated datasets w.r.t both motion diversity and consistency.

\subsection{Ablation study}

\noindent\textbf{Motion styles and MoStAtt.}
We first investigate the effect of motion styles and how MoStAtt mechanism contributes to the performance.
We compare with the baseline method StyleGAN-V~\cite{skorokhodov2022stylegan} which does not have time-variant motion styles, i.e. $K=0$. 
We also conduct experiments using different numbers of motion styles, with $K$ selecting from $[1,6,8,10]$. Note that when $K=1$, which means only one motion style will attend in the modulation matrix instead of adaptively attentive modulation what MoStAtt is designed for, hence we regard it as w/o MoStAtt.

Table \ref{tab:k} shows the performance of our method with only one motion style ($K=1$) w/o using MoStAtt drops compare to the baseline, but is largely boosted when multiple motion styles are used along with the MoStAtt mechanism.  A small number of motion styles (e.g. $K=6$) can not fully capture various motion patterns; meanwhile, too many motion styles ($K=10$) could not improve performance but lead to ineffective computation. In addition, redundant motion styles would not contribute to motion synthesis because they will potentially receive small attention scores. Our method achieves the best performance when $K=8$.

\begin{table}[!htbp]
    \centering
\begin{adjustbox}{width=1\linewidth,center}
\begin{tabular}{lcccccccc}
\toprule  \multirow{2}{*}{ \textbf{Number of Motion Styles} } & \multicolumn{2}{c}{ \textbf{FaceForensics $\mathbf{256^2}$} } & \multicolumn{2}{c}{ \textbf{RainbowJelly $\mathbf{256^2}$} } & \multicolumn{2}{c}{ \textbf{CelebV-HQ $\mathbf{256^2}$} }
 \cr & $\rm{FVD_{16}}\downarrow$ & $\rm{FVD_{128}}\downarrow$ & $\rm{FVD_{16}}\downarrow$ & $\rm{FVD_{128}}\downarrow$ & $\rm{FVD_{16}}\downarrow$ & $\rm{FVD_{128}}\downarrow$ \\ 
 \midrule
K=0 (StyleGAN-V~\cite{skorokhodov2022stylegan}) &47.4&89.3&195.4&262.5&68.0&158.6 \\
K=1 (w/o MoStAtt) & 50.9 & 104.6 & 248.7 & 122.1 & 68.4 &  172.0 \\
K=6 & 46.3 & 83.9 & 96.1 & 89.9  & 60.8 & 143.5 \\
K=8 (default) & \textbf{39.7} & \textbf{72.6} & \textbf{70.1} & \textbf{74.3} & 56.1 & 132.1  \\
K=10 & 40.4 & 79.8 & 77.6 & 77.9 & \textbf{54.8} & \textbf{125.8} \\
\bottomrule
\end{tabular}
\end{adjustbox}
\caption{Effect of MoStAtt and different numbers of motion styles. Note that $K=0$ is StyleGAN-V~\cite{skorokhodov2022stylegan} baseline for comparison and $K=1$ indicates only single motion style will attend in modulation matrix and thus regarded as w/o MoStAtt.}
\label{tab:k}
\end{table}

\noindent \textbf{Modulation Order.} In our MoStGAN-V, two different styles are used for weight modulation, i.e., content style and motion style. An important design choice is the order of style-based weight modulation. We attempt two  strategies:

\noindent \textbf{i)} This is the default setting of our method. We first modulate weights of each synthesis layer with the content style $s$ and then use $K$ motion styles $\{\tilde{m}_t^k\}_{k=1,2,...,K}$ with MoStAtt to module the weights as Section~\ref{sec:attn} introduced.

\noindent \textbf{ii)} We fist modulate the weights of each synthesis layer with $K$ motion styles $\{\tilde{m}_t^k\}_{k=1,2,...,K}$ with MoStAtt mechanism (use $\mathbf{W}_G$ instead of $\mathbf{W}_G^{'}$ in Eq. (\ref{eq:attn1})) and then, similar to original StyleGAN2~\cite{karras2020training}, use content style $s$ for weight modulation.

Intuitively, the first strategy regards the content-style-modulated weights as the whole context of a video. Then the following MoStAtt assigns significance to motion styles that correspond to possible motions belonging to the current frame given the whole context. Table \ref{tab:modulate} shows the first strategy has significantly better performance than the one that switches the modulation order. The results verify the advantage of our design choice that first modules the weights using a time-agnostic style at a global perspective and then time-sensitive motion styles at a local perspective.

\begin{table}[!htbp]
    \centering
\begin{adjustbox}{width=1\linewidth,center}
\begin{tabular}{lcccccccc}
\toprule  \multirow{2}{*}{ \textbf{Modulation} } & \multicolumn{2}{c}{ \textbf{FaceForensics $\mathbf{256^2}$} } & \multicolumn{2}{c}{ \textbf{RainbowJelly $\mathbf{256^2}$} } & \multicolumn{2}{c}{ \textbf{CelebV-HQ $\mathbf{256^2}$} }
 \cr & $\rm{FVD_{16}}\downarrow$ & $\rm{FVD_{128}}\downarrow$ & $\rm{FVD_{16}}\downarrow$ & $\rm{FVD_{128}}\downarrow$ & $\rm{FVD_{16}}\downarrow$ & $\rm{FVD_{128}}\downarrow$ \\ 
 \midrule
strategy i (default) & 39.76 & 72.64 & 70.10 & 74.39 &56.17 &132.14 \\
strategy ii & 65.07 & 150.74 &124.18 &121.83 & 76.85 & 209.90 \\
\bottomrule
\end{tabular}
\end{adjustbox}
\caption{Different weight modulation strategies.}
\label{tab:modulate}
\end{table}

\noindent \textbf{Motion Difference Capture.}
We further investigate the effect of using motion difference as discriminator inputs to encourage motion consistency among frames. As Table \ref{tab:dim_m} shows, adding motion difference as additional discriminator inputs does not improve the performance of StyleGAN-V model, while significantly boost the performance of our MoStGAN-V model. This indicates our MoStAtt-augmented generator can produce realistic enough motions and fight against the discriminator, making it more capable of capturing motion artifacts.

\begin{table}[!htbp]
    \centering
\begin{adjustbox}{width=0.9\linewidth,center}
\begin{tabular}{lcccccc}
\toprule  \multirow{2}{*}{ \textbf{Methods} } & \multicolumn{2}{c}{ \textbf{FaceForensics $\mathbf{256^2}$} } & \multicolumn{2}{c}{ \textbf{CelebV-HQ $\mathbf{256^2}$} }
 \cr & $\rm{FVD_{16}}\downarrow$ & $\rm{FVD_{128}}\downarrow$ & $\rm{FVD_{16}}\downarrow$ & $\rm{FVD_{128}}\downarrow$ \\ 
 \midrule

StyleGAN-V~\cite{skorokhodov2022stylegan} & 47.4 & 89.3 & 68.0 & 158.6  \\
$\,\,\,\,\,$w/ motion-diff & 58.1 & 140.6 & 88.4 & 204.2  \\
MoStGAN-V (ours) & 49.1 & 117.8 & 61.0 & 138.6 \\
$\,\,\,\,\,$w/ motion-diff (default)  & \textbf{39.7} & \textbf{72.6} & \textbf{56.1} & \textbf{132.1} \\

\bottomrule
\end{tabular}
\end{adjustbox}
\caption{Motion difference capture. We conduct experiments w and w/o taking motion difference as discriminator inputs on StyleGAN-V~\cite{skorokhodov2022stylegan} and our model.}
\label{tab:dim_m}
\end{table}



\subsection{Motion Diversity} 
\begin{figure}[!htbp]
\begin{adjustbox}{width=\linewidth,center}
    \centering
    \includegraphics{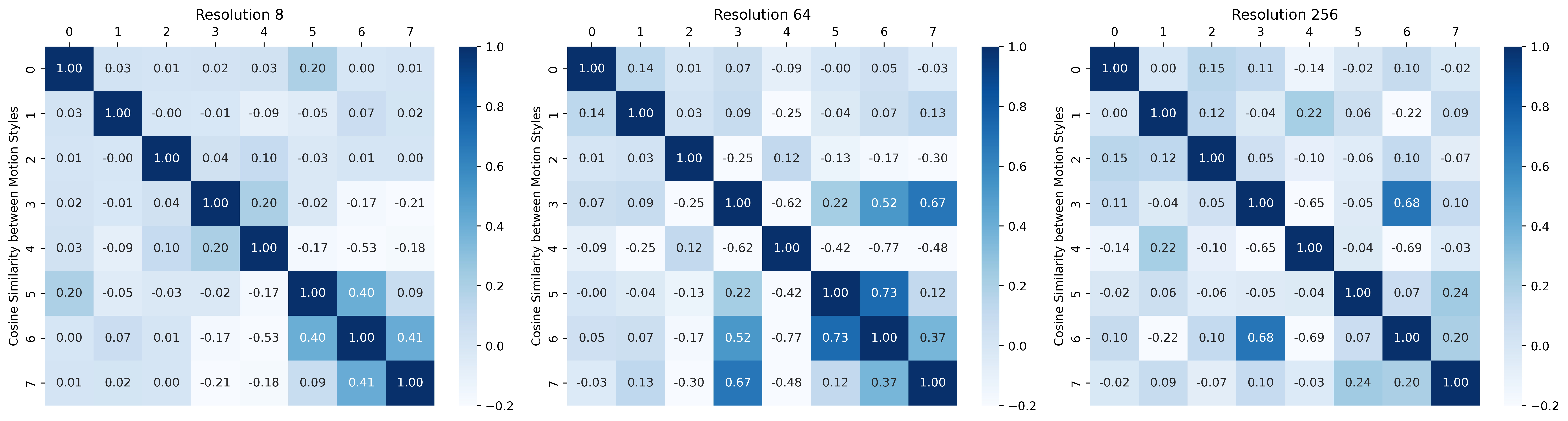}
\end{adjustbox}
\caption{Cosine similarities between different motion styles at the first \texttt{ModConv2d} layer of synthesis blocks at resolutions of $8^2$, $64^2$ and $256^2$ respectively on CelebV-HQ $256^2$~\cite{zhu2022celebv}.}
\label{fig:sim}
\end{figure}

\begin{figure}[!htbp]
\begin{adjustbox}{width=1.0\linewidth,center}
\centering
\includegraphics{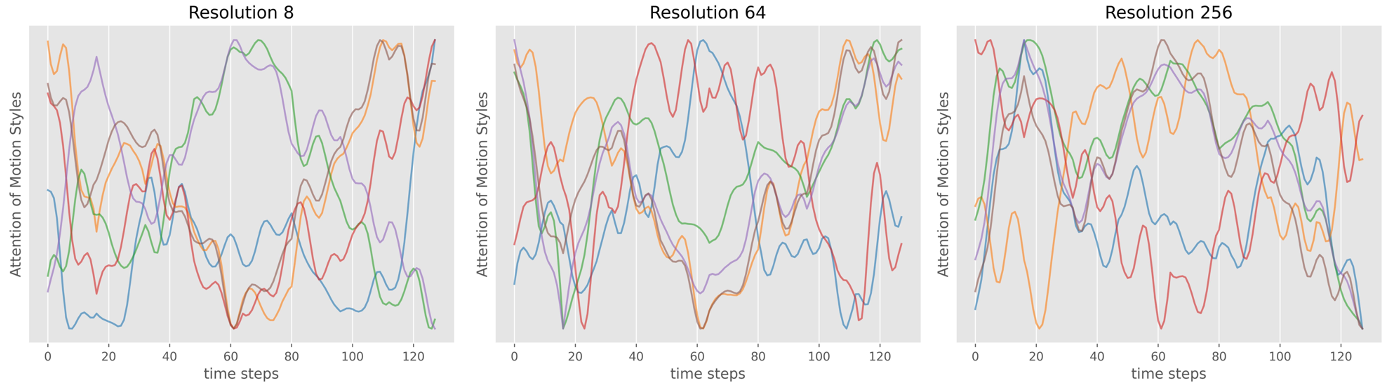}
\end{adjustbox}
\caption{Attention scores for different motion styles in $\mathbb{R}^{c_{out}}$ over time at the first \texttt{ModConv2d} layer of synthesis blocks at resolutions of $8^2$, $64^2$ and $256^2$ respectively on CelebV-HQ $256^2$~\cite{zhu2022celebv}.}
\label{fig:att_score}
\end{figure}

The motion styles are introduced to model different motion patterns in the generated video. Therefore, we quantitatively analyze a) the diversity of the motion styles; b) how the attention scores change over time, to interpret the behavior of motion styles  in $\mathcal{S}$ space. Figure~\ref{fig:sim} shows cosine similarities between $K$ motion styles in different scales, i.e., the first \texttt{ModConv2d} layer of synthesis blocks at resolutions of $8^2$,$64^2$ and $256^2$ respectively. Figure~\ref{fig:sim} shows different motions styles have small similarity scores, which indicates that they are encouraged to be disentangled from each other and model different motion patterns. Figure~\ref{fig:att_score} shows attention scores assigned by deconvolutional filters for different motion styles among time steps. From this figure, we can tell that motion styles are dynamically attending weight modulation matrix to model different motion patterns.

\section{Conclusion}
In this paper, we rethink a single time-agnostic latent vector of traditional style-based method is insufficient to model temporal motion patterns and thus introduce time-dependent motion styles to facilitate video generation with vivid dynamics. A motion style attention modulation mechanism is further designed to dynamically assign different motion styles to the current frame, which will be used to modulate the deconvolution filter weights in the target synthesis layer. Our method achieves state-of-the-art performance on four video generation benchmarks and the MoStAtt-enhanced generator can produce better qualitative results by generating diverse and temporal coherent motions.

\section*{Acknowledgment}
This work was supported by KAUST BAS/1/1685-01-01, SDAIA-KAUST Center of Excellence in Data Science and Artificial Intelligence.

{\small
\bibliographystyle{ieee_fullname}
\bibliography{egbib}
}

\appendix

\newpage
\clearpage

\section*{Appendix}

\section{Experiments}

\subsection{Motion Diversity Loss}
In Section 3.4 we introduce motion diversity loss $L_{div}$ to encourage motion styles to be disentangled from each other for modeling various motions pattern. Table~\ref{tab:div} shows the ablations for w/ or w/o $L_{div}$. When adding $L_{div}$, FaceForensics and CelebV-HQ gain a little improvement with $FVD_{128}$ from 79.8 to 72.6 and 166.1 to 132.1 respectively, however, for SkyTimelapse the performance drops a little. One possible reason is that SkyTimelapse contains only simple motions like clouds slowly moving in one direction.
\begin{table}[!htbp]
    \centering
\begin{adjustbox}{width=0.9\linewidth,center}
\begin{tabular}{lcccccccc}
\toprule  \multirow{2}{*}{ \textbf{Method} } & \multicolumn{2}{c}{ \textbf{FaceForensics $\mathbf{256^2}$} } & \multicolumn{2}{c}{ \textbf{SkyTimelapse $\mathbf{256^2}$} } & \multicolumn{2}{c}{ \textbf{CelebV-HQ $\mathbf{256^2}$} }
 \cr & $\rm{FVD_{16}}\downarrow$ & $\rm{FVD_{128}}\downarrow$ & $\rm{FVD_{16}}\downarrow$ & $\rm{FVD_{128}}\downarrow$ & $\rm{FVD_{16}}\downarrow$ & $\rm{FVD_{128}}\downarrow$ \\ 
 \midrule
MoStGAN-V(ours) & 39.7 & 72.6 & 65.3 & 162.4 & 56.1 & 132.1 \\
\,\,\,\, w/o $L_{div}$ &40.4&79.8&65.8&143.2 & 58.4& 166.1\\
\bottomrule
\end{tabular}
\end{adjustbox}
\caption{Ablations on different rank.}
\label{tab:div}
\end{table}

\subsection{Magnitude of Latent Motion Vectors} 
In addition to content latent vector $w$ in original StyleGAN2~\cite{karras2020training}, we propose to separate motion style from original content vector for better motion synthesis controlment. Since both will later be transformed into style parameters to modulate the weights of each Synthesis Layer, here we want to explore the importance of each contribution by changing the 
dimension of each motion latent vector $\{m_t^k\}_{k=1,2,...,K}$ after the Motion Network $\sf{F_m}$ and remaining the dimension of latent vector $w$ as $512$, same as the original model.

\begin{table}[!htbp]
    \centering
\begin{adjustbox}{width=0.9\linewidth,center}
\begin{tabular}{lcccccccc}
\toprule  \multirow{2}{*}{ \textbf{Dimension} } & \multicolumn{2}{c}{ \textbf{FaceForensics $\mathbf{256^2}$} } & \multicolumn{2}{c}{ \textbf{SkyTimelapse $\mathbf{256^2}$} } & \multicolumn{2}{c}{ \textbf{CelebV-HQ $\mathbf{256^2}$} }
 \cr & $\rm{FVD_{16}}\downarrow$ & $\rm{FVD_{128}}\downarrow$ & $\rm{FVD_{16}}\downarrow$ & $\rm{FVD_{128}}\downarrow$ & $\rm{FVD_{16}}\downarrow$ & $\rm{FVD_{128}}\downarrow$ \\ 
 \midrule
d=64 & 72.5 & 153.6 & 78.6 & 162.3 & 76.3 & 158.0 \\
d=128 (default) & 39.7 & 72.6 & 65.3 & 162.4 & 56.1 & 132.1\\
d=256 & 52.4 & 125.9 & 74.7 & 161.3 & 64.9 & 150.7 \\
\bottomrule
\end{tabular}
\end{adjustbox}
\caption{Different dimensions of latent motion vector $m_t^k$.}
\label{tab:dim}
\end{table}

\subsection{Number of rank}

Table~\ref{tab:rank} shows the result of different rank $R$ for low-rank factorization. Increasing the rank could not improve the performance much but introduce more parameters instead, thus we set $R=1$ as default.

\begin{table}[!htbp]
    \centering
\begin{adjustbox}{width=0.9\linewidth,center}
\begin{tabular}{lcccccccc}
\toprule  \multirow{2}{*}{ \textbf{Rank} } & \multicolumn{2}{c}{ \textbf{FaceForensics $\mathbf{256^2}$} } & \multicolumn{2}{c}{ \textbf{SkyTimelapse $\mathbf{256^2}$} } & \multicolumn{2}{c}{ \textbf{CelebV-HQ $\mathbf{256^2}$} }
 \cr & $\rm{FVD_{16}}\downarrow$ & $\rm{FVD_{128}}\downarrow$ & $\rm{FVD_{16}}\downarrow$ & $\rm{FVD_{128}}\downarrow$ & $\rm{FVD_{16}}\downarrow$ & $\rm{FVD_{128}}\downarrow$ \\ 
 \midrule
R=1(default) & 39.7 & 72.6 & 65.3 & 162.4 & 56.1 & 132.1 \\
R=3 & 40.6 & 70.4 & 65.7 & 154.1 & 56.6 & 136.9 \\
R=5 & 39.5 & 74.9 & 75.5 & 163.2 & 53.8 & 152.8 \\
R=10& 37.6 & 64.2 & 83.3 & 174.0 & 55.7 & 157.6 \\
\bottomrule
\end{tabular}
\end{adjustbox}
\caption{Ablations on different rank.}
\label{tab:rank}
\end{table}

\subsection{Generating videos with diverse background}

UCF101~\cite{soomro2012ucf101} is a challenging dataset with large variations, e.g., actions, camera motion, object appearance. As Table~\ref{tab:ucf} shows, all models struggle to achieve a satisfying FVD and our method performs slightly better.
In addition, we also conducted experiments on Horseback~\cite{brooks2022generating} (resized to $256^2$) with moving camera. Compared with StyleGAN-V~\cite{skorokhodov2022stylegan} with $\rm{FVD_{16}}$ performance of 168.14, our model achieves better result of 146.65.

\begin{table}[!htbp]
    \renewcommand\arraystretch{1.2}
\begin{minipage}{.47\linewidth}
  \centering
    \begin{adjustbox}{width=\linewidth,center}
      \begin{tabular}{cccccc}
\toprule   \textbf{Methods} & $\rm{FVD_{16}}\downarrow$ & $\rm{FVD_{128}}\downarrow$\\ 
 \midrule
MoCoGAN-HD~\cite{tian2021mocogan} & 1729.6 & 2606.5 \\
DIGAN~\cite{yu2022digan} & 1630.2 & 2293.7 \\
StyleGAN-V~\cite{skorokhodov2022stylegan} & 1431.0 & 1773.4 \\
MoStGAN-V (ours) & 1380.3  & 1695.6 \\
\bottomrule
\end{tabular}
\end{adjustbox}
\caption{Comparison on UCF101~\cite{soomro2012ucf101} dataset.}
\label{tab:ucf}
\end{minipage}
\hfill  
\begin{minipage}{.48\linewidth}
  \centering
\begin{adjustbox}{width=\linewidth,center}
      \begin{tabular}{cccccc}
\toprule  \multirow{1}{*}{ \textbf{Methods} } & \multicolumn{1}{c}{ $\rm{FVD_{16}}\downarrow$  } & \multicolumn{1}{c}{ $\rm{\# Params}\downarrow$  }\\ 
 \midrule

StyleGAN-V$_{text}$~\cite{skorokhodov2022stylegan} & 191.1 & 58M  \\
MUGEN~\cite{hayes2022mugen} & 112.7 & 120M \\
TATS~\cite{ge2022long} & 89.3 & 562M \\
MoStGAN-V$_{text}$ (ours) & 129.8 & 66M \\

\bottomrule
\end{tabular}
\end{adjustbox}
\caption{Comparison on text-conditional MUGEN~\cite{hayes2022mugen}.}
\label{tab:text}
\end{minipage}
\end{table}

\subsection{Extending to text-conditional video generation}

We extended our approach to a text-conditional model (denoted as MoStGAN-V$_{text}$) with two granularity information as inputs. In sentence level, we concatenate noise vector $z^c$ with text embedding and pass it through Mapping Network $\sf{F_c}$ to obtain latent content vector $w$; while in word level, we extract words embeddings for latent motion vectors $m_t^k$ (since MUGEN~\cite{hayes2022mugen} dataset contains frame-level fine-grained information). We enable text-conditional StyleGAN-V~\cite{skorokhodov2022stylegan} (denoted as StyleGAN-V$_{text}$) by concatenating noise vector $z^c$ with text embedding.
Table~\ref{tab:text} shows our method is superior to StyleGAN-V~\cite{skorokhodov2022stylegan}. Compared to two-stage models TATS~\cite{ge2022long} and MUGEN~\cite{hayes2022mugen}, i.e., firstly train a VQGAN~\cite{esser2021taming} and a Transformer in the second stage, our model achieves comparable result with efficiency. This is a toy model to show our method is extendable for conditional generation and we leave further improvement for future exploration.

\section{Properties}

\subsection{Motion Style Interpretation}
\begin{figure}[!htbp]
\begin{adjustbox}{width=\linewidth,center}
    \centering
    \includegraphics{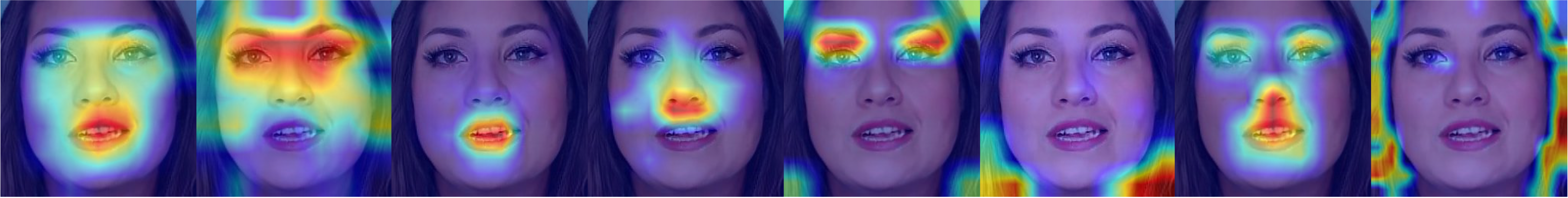}
\end{adjustbox}
\caption{Each grid represents for implementing attention map on top of generated frames for different motion styles at the same time step.}
\label{fig:interp}
\end{figure}

We investigate how different motion styles exert influence to generated frames.
During inference, we analyze how different motion styles response to the output feature by calculating attention map with dimesionality $K\times H\times W$ between attention matrix $\mathbf{A_t} \in \mathbb{R}^{K\times c_{out}}$ and output feature of the \texttt{Modconv2d} $ F_t \in \mathbb{R}^{c_{out} \times H \times W}$ in time step $t$, where $K$ corresponds to $K$ motion styles. For better visualization, we calculate attention map in last \texttt{Modconv2d} layer of synthesis block $256^2$ and add it on top of the final generated frames.
Note that we do not focus on disentangled representations, i.e., considering a latent representation to be perfectly disentangled if each latent dimension controls a single visual attribute~\cite{ridgeway2018learning,eastwood2018framework}, but only want to explore the influence of our proposed motion styles. Figure~\ref{fig:interp} shows that attention map for each motion style tends to focus on different regions correspond to different motion patterns, e.g., blinking eyes or open mouth.

\subsection{Motion Content Decomposition}
\begin{figure}[!htbp]
\begin{adjustbox}{width=0.9\linewidth,center}
    \centering
    \includegraphics{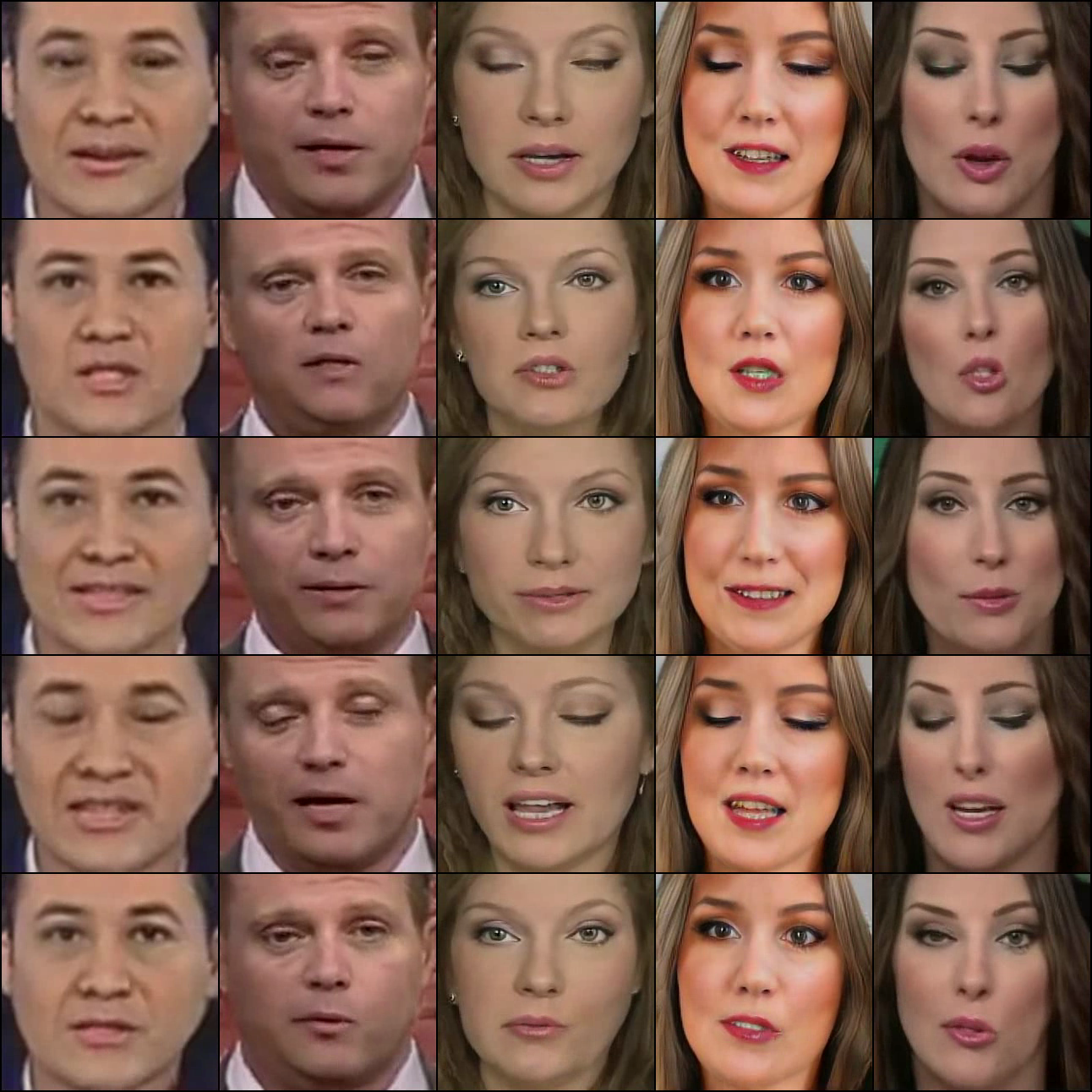}
\end{adjustbox}
\caption{Each row sample from same motion noise $z_{t_0}^m,...,z_{t_n}^m$ while each column starts from same content noise $z^c$ and each grid presents different videos at same time step.}
\label{fig:democo}
\end{figure}

For better observation, we control univariate to show how additional motion style achieves better motion synthesis. Each column of Figure~\ref{fig:democo} shows diverse motions originates the same content $z^c$ that controls the appearance variances, while each row shares same motion noises $z_{t_0}^m,...,z_{t_n}^m$ with variant identities. Our model perform better than StyleGan-V that can ensure sampling from same motion codes will lead to consistent motions (see each rows, e.g., blink eyes, head posture, mouth opening size and direction.)
Please watch grid videos in our anonymous webpage for better observation.

\section{Efficiency of low-rank factorization}

The hypernetworks are introduced to produce modulation matrix for weights of each \texttt{ModConv2d} layer. If the hidden layer size of hypernetwork is of dimensionality $d_h=128$ and the deconvolutional weight tensor is the size of $d_o = c_{out} \times c_{in} \times k_h \times k_w=512\times 512 \times 3 \times 3 \approx 2.4$ million, where $c_{out}$, $c_{in}$ is the dimensionality of output and input channel, and $k_h, k_w$ are the kernel size of deconvolutional filters. Then the output weight matrix in the hypernetwork will be of size $d_h \times d_o \approx  0.3$ billion, which is memory-intensive for each layer. To alleviate this issue, we leverage low-rank factorization for the modulation matrix. The modulation tensor for each motion style of each \texttt{ModConv2d} layer within our proposed MoStAtt as Section 3.3 mentioned should have the dimensionality of $c_{in} \times k_h \times k_w$. Here we omit the rank $R$ for simplicity. To make a light-wise architecture in a parameter-effective manner, we decompose the modulation matrix with low-rank factorization. Therefore, the output dimensionality of hypernetwork will be reduced from $c_{in} \times k_h \times k_w$ to $c_{in} + k_h + k_w$. Therefore, by using low-rank factorization, the number of parameters of hypernetwork in current \texttt{ModConv2d} layers will be reduced from 0.3 billion to $d_h \times (c_{in} + k_h + k_w) \approx 0.07$ million, which largely reduces the computation burden.

\section{Human Evaluation}
\begin{figure}[!t]
\begin{adjustbox}{width=0.9\linewidth,center}
    \centering
    \includegraphics{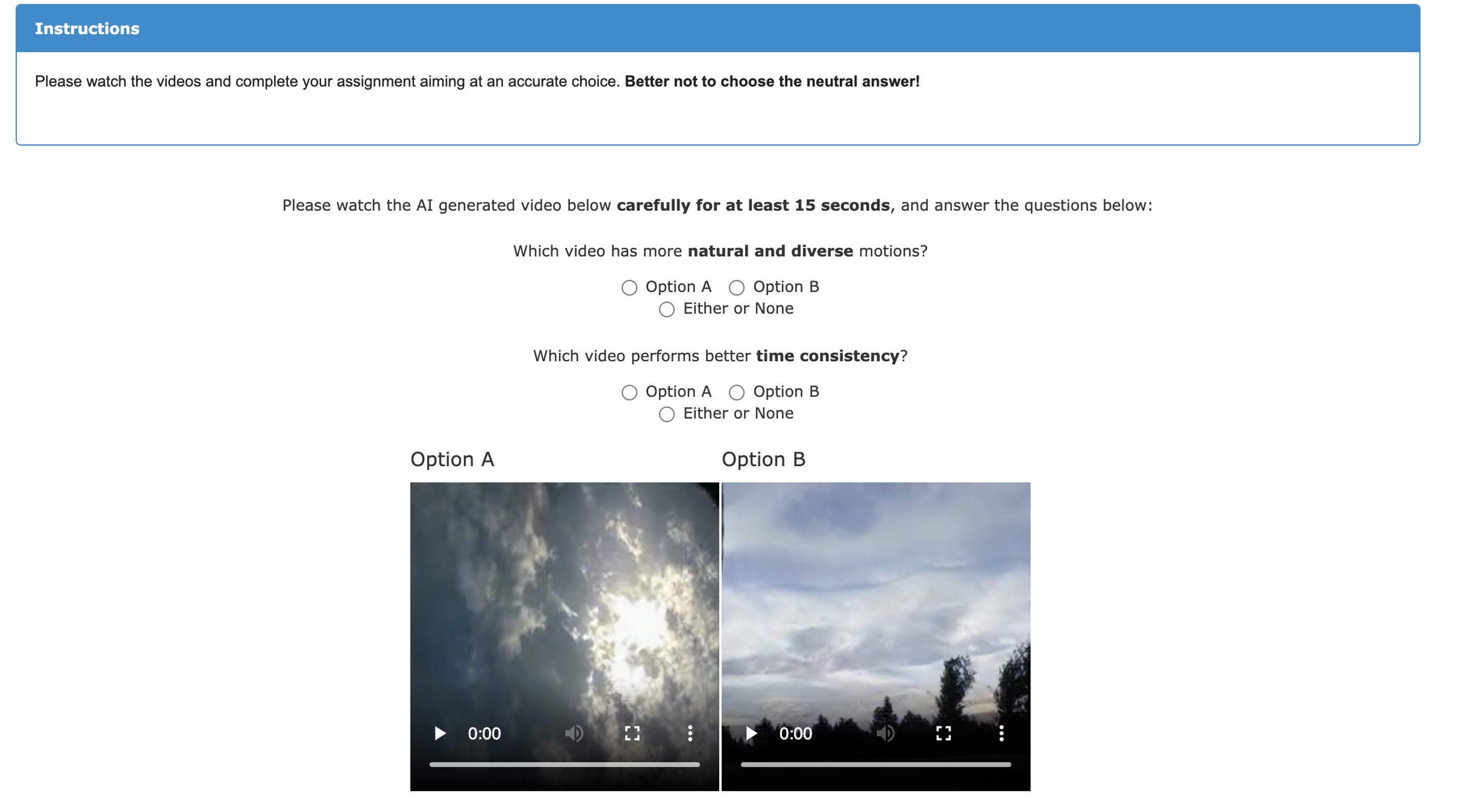}
\end{adjustbox}
\caption{Webpage interface for human evaluation.}
\label{fig:page}
\end{figure}

We conducted a human evaluation on Amazon Mechanical Turk to assess videos generated by our method in comparison to StyleGAN-V~\cite{skorokhodov2022stylegan} w.r.t motion diversity as well as temporal consistency. We provide 100 pairs of videos each for three datasets FaceForensics, RainbowJelly and SkyTimelapse and pairwise compare 2 random videos in random order from either our method or StyleGAN-V~\cite{skorokhodov2022stylegan} trained on the same dataset. And we provide the Mturker with two questions: \textit{'Which video has more natural and diverse motions?'} and \textit{'Which video performs better time consistency?'}, based on which they will choose their preference and if it is hard to decide a better one, we also provide a neutral option. Figure~\ref{fig:page} shows how the interface guide users to conduct the evaluation. Each video pair is assigned to 5 unique workers, resulting in 500 responses for each dataset. The average time per assignment is 10 minutes and 37 seconds.

\section{Datasets details}
\textbf{FaceForensics~\cite{rossler2018faceforensics}} FaceForensics is a video dataset where all videos are downloaded from Youtube and cut down to short continuous clips that contain mostly frontal faces.
We follow  previous work\footnote{\url{https://github.com/universome/stylegan-v/blob/master/src/scripts/preprocess_ffs.py}} to extract face crops for this talking head datasets, which crops each frame independently and somehow makes unstable shaking.

\textbf{CelebV-HQ~\cite{zhu2022celebv}} CelebV-HQ is a large-scale high-quality video dataseet with rich celebrity identities and actions. We preprocess the dataset with official link\footnote{\url{https://github.com/CelebV-HQ/CelebV-HQ/blob/main/download_and_process.py}}.

\textbf{SkyTimelapse~\cite{xiong2018learning}}
SkyTimelapse contains slow motions of sky changing under different time and weather conditions. We use the released 2,377 videos from official dataset although they~\cite{xiong2018learning} claims that 5, 000 videos are collected\footnote{\url{https://github.com/weixiong-ur/mdgan}}.

\textbf{RainbowJelly~\cite{skorokhodov2022stylegan}} 
RainbowJelly dataset is an 8-hour-long movie of 4K resolution underwater video with colorful jellyfishes. We follow previous work\footnote{\url{https://github.com/universome/stylegan-v/blob/master/src/scripts/convert_video_to_dataset.py}} and divide it into clips of 512 frames each. We use it as a video generation benchmark to evaluate the effect of our MoStAtt since it contains complex hierarchical motions.

For all the datasets, we center crop to $256^2$ resolution and use train split if available to train all the models. All the dataset are 25 fps except RainbowJelly is 30 fps.

\end{document}